\documentclass[10pt,a4paper,twocolumn]{article}
\usepackage[utf8]{inputenc}
\usepackage{amsmath}
\usepackage{amsfonts}
\usepackage{amssymb}
\usepackage{graphicx}
\usepackage{float}
\usepackage{booktabs}
\usepackage{multirow}
\usepackage[linesnumbered,ruled,vlined]{algorithm2e}
\usepackage{changes}  
\usepackage{array}
\title{Efficient and Secure Federated Learning for Financial Applications}

\author{Tao Liu, Business School, China University of Political Science and Law
\and 
Zhi Wang, The School of Software, Xi’an Jiaotong University 
\and 
 Hui He~\footnote{huihe@xjtu.edu.cn}, Wei Shi, Liangliang Lin, Wei Shi, Ran An, \\ The School of Computer Science and Technology, Xi’an Jiaotong University
 \and 
 Chenhao Li, Ant Rongxin (Chengdu) Network Technology Co., Ltd}

\date{March, $1^{st}$, 2023}

\begin{document}
\maketitle

\section*{Absctract}
The conventional machine learning (ML) and deep learning approaches need to share customers' sensitive information with an external credit bureau to generate a prediction model that opens the door to privacy leakage. This leakage risk makes financial companies face an enormous challenge in their cooperation. Federated learning is a machine learning setting that can protect data privacy, but the high communication cost is often the bottleneck of the federated systems, especially for large neural networks. Limiting the number and size of communications is necessary for the practical training of large neural structures. Gradient sparsification has received increasing attention as a method to reduce communication cost, which only updates significant gradients and accumulates insignificant gradients locally. However, the secure aggregation framework cannot directly use gradient sparsification. This article proposes two sparsification methods to reduce communication cost in federated learning. One is a time-varying hierarchical sparsification method for model parameter update, which solves the problem of maintaining model accuracy after high ratio sparsity. It can significantly reduce the cost of a single communication. The other is to apply the sparsification method to the secure aggregation framework. We sparse the encryption mask matrix to reduce the cost of communication while protecting privacy. Experiments show that under different Non-IID experiment settings, our method can reduce the upload communication cost to about 2.9\% to 18.9\% of the conventional federated learning algorithm when the sparse rate is 0.01.

\textbf{Key word:} Federated learning; gradient sparsification; secure aggregation; communication cost optimize

\section{Introduction}
Modern financial firms routinely need to conduct analysis of large data sets stored across multiple servers or devices. A typical response is to combine those data sets into a single central database, but this approach introduces a number of privacy challenges: The institution may not have appropriate authority or permission to transfer locally stored information, the owner of the data may not want it shared, and centralization of the data may worsen the potential consequences of a data breach.
Federated Learning (FL)\cite{mcmahan2017communication,DBLP:journals/spm/LiSTS20} allows users to form a joint training to obtain a global model without explicitly sharing the local private data, so as to effectively solve the problems of data privacy and security. However, the communication cost is often the bottleneck of FL, because a large number of devices send their local updates to the central server\cite{DBLP:journals/ftml/KairouzMABBBBCC21}. The iteration of the aggregation model relies on high-frequency communication between a large number of clients and a single server, and there is an upper limit on the bandwidth resources of them. How to use the limited computing and communication resources to obtain the optimal learning performance is an urgent problem to be solved in federated learning\cite{8664630}. Compared with distributed learning, the communication connection between each computing node in FL is relatively fragile. In this case, the communication volume of each aggregation update becomes a key factor restricting the efficiency of FL. The larger the volume of the parameter update vector transmitted each time\cite{DBLP:journals/corr/abs-2011-07429}, the longer the communication time required, and the average time between two parameter aggregation operations on the server will be prolonged. At the same time, the longer it takes to send the parameter update vector, the probability of facing abnormal situations such as disconnection and network fluctuation will increase accordingly, so the robustness requirements for the FL algorithm will also become higher. Therefore, for a FL algorithm with high communication efficiency,  these updates must be sent in a compressed and infrequent manner.

Gradient sparsification\cite{huang2017densely} is a common method of gradient compression to reduce the communication cost. The main idea of Gradient sparsification is that after each local iterative calculation process, due to the difference in the convergence speed of different parameters of the model itself, in the process of one iteration, some parameters may have small iteration steps, and others have larger ones. In each communication, the data volumes are the same no matter whether they are consumed by transmitting the change value of a small or large parameter, but the parameter update with a large change has a greater impact on the performance of the updated model. Therefore, when transmitting parameter updates, only the parameters with large changes can be transmitted, and the parameters with small changes can be temporarily stored in local accumulation\cite{sattler2019sparse}, and wait until the accumulated value reaches a certain range before transmitting to the server. The advantage is that each communication is transmitting more important parameter updates with less communication volume.

However the current sparse method still has some unreasonable points. For a deep neural network, its powerful learning ability comes from the deeper structure of the network, and each layer has its own specific function. For convolutional neural networks (CNN) suitable for image recognition, studies have shown that each layer of CNN corresponds to different feature extraction capabilities. As the number of network layers increases, the extracted features become more and more abstract. The parameters of each layer of a deep neural network have their own characteristics. Specifically in terms of numerical value, the network parameters of different layers have orders of magnitude difference. So after the model parameters are directly flattened into a one-dimensional vector, the network parameters with smaller values will be covered by the larger parameter values in the process of sparseness and will not be uploaded, but this does not mean that the parameter updating of this part is not important. The cumulative residual of gradients not uploaded by themselves have a certain delay. If there are too many cumulative rounds, the iteration direction it represents is no longer the descent direction required by the current model parameters\cite{bernstein2018signsgd}. If these parameters are always covered by other larger parameters and are treated as residuals in each gradient sparsification process, the loss caused by sparsification will be correspondingly larger.

Another key consideration for FL system is to preserve the privacy of the users. Unfortunately, FL has been demonstrated to leak sensitive information\cite{DBLP:journals/corr/abs-1812-00984,carlini2019secret,melis2019exploiting} by sharing intermediate model updates. To combat it, FL often couple together Differential Privacy (DP)\cite{dwork2014algorithmic} model training as well as Homomorphic Encryption (HE)\cite{acar2018survey} or Secure Multiparty Computation (SMC)\cite{evans2017pragmatic} secure aggregation techniques to ensure that transmitted model information is privacy preserving. HE allows aggregations to be performed on encrypted data\cite{halevi2011secure,leontiadis2014private}, but the main problem with this method is that performing computations in the encrypted domain is computationally expensive. SMC can be used to securely aggregate local model updates without leaking them, but it also faces the communication bottleneck of traditional distributed training. Bonawitz et al. (2017) proposed Secure Aggregation (SA)\cite{bonawitz2017practical}, which allows a server to compute the sum of large, user-held data vectors from mobile devices in a secure manner without learning each user's individual contribution. However, SA requires significant amounts of additional resources on communication and computing for protecting privacy.

Today, the challenge of FL is to build a protocol that provides privacy protection, which also has computation and communication efficiency. All of them will not significantly affect the accuracy. {In this paper, we propose an efficient and secure FL framework based on gradient and mask sparsification, which consists of two parts: time-varying hierarchical gradient sparsification and encryption mask sparsification. The time-varying hierarchical gradient sparsification algorithm can balance the numerical size difference of network parameters at different layers, so as to reduce the loss caused by the sparsification process. In addition, we sparsify the encryption mask in the secure aggregation framework to reduce the total amount of data sent and improve the communication efficiency while ensuring security.} Our contributions can be summarized as following:

\begin{enumerate}
\item We propose a time-varying hierarchical sparse method for model parameter update. This method can greatly reduce the cost of a single communication and have a smaller loss of model performance.
\item We design a sparse mask matrix method of calculating encryption mask matrix with zero local value, which reduces the amount of data transmitted by the secure aggregation framework during communication.
\item Our experiments on MNIST, Fashion-MNIST and CIFAR-10 prove that our method can reduce the upload communication cost to about 2.9\% to 18.9\% of the conventional FL algorithm when the sparsity rate is 0.01.
\end{enumerate}

{The structure of this work is arranged in the following. Section 2 introduces related works and up-to-date scholar leading edge upon gradient sparsification and secure aggregation. In Section 3, the efficient and secure federated learning framework is presented, including time-varying hierarchical gradient sparsification and encryption mask sparsification. Section 4 describes experimental settings and results on the methodology. Finally, Section 5 presents our conclusions about the proposed framework.}

\section{Related Work}
\subsection{Gradient Sparsification}
In the communication optimization of FL, researchers have proposed many methods to solve the communication bottleneck problem. From the system perspective, Zinkevich et al. (2010)\cite{he2016deep} proposed an asynchronous update system, in which each node can update the model independently without being restrained or affected by other nodes with slower update speed. From the specific perspective of communication transmission, reducing the traffic of each round is a method to directly reduce the communication time. Gradient sparsification is a common gradient compression method to reduce the communication burden\cite{NEURIPS2018_17326d10}.

Strom et al. (2015)\cite{strom2015scalable} proposed a gradient dropping method, which only sends gradients larger than a predefined constant threshold, {and all other gradients are accumulated in the locally saved residuals and not sent temporarily. This method can achieve up to 3 orders of magnitude compression ratio of uploaded data. However, in the actual training task, it is difficult to choose the appropriate value for the threshold. Due to the different thresholds for different tasks and different models, artificial scheduling will introduce a lot of experimental costs. In order to overcome this problem, Dryden et al. (2016)\cite{7835789} made some improvements on Strom's sparsification method. Compared with the fixed threshold of Storm method, the sparsity rate \(s\) is fixed. For example, when \(s\) = 0.001, it means that only the top 0.1\% of the gradient update vector is transmitted, and other parts of the gradient update are kept locally for the time being. Their experiments show that with a sparsity rate of \(s\) = 0.001, their method only slightly reduces the convergence speed and the final accuracy of the speed training model.}

Lei Ba et al. (2016)\cite{lei2016layer} believe that normalization on each layer is necessary to ensure the convergence of gradient dropping. Lin et al. (2018)\cite{DBLP:conf/iclr/LinHM0D18} proposed the gradient sparsification algorithm of deep gradient compression (DGC) in order to solve the gradient redundancy problem in distributed training. The algorithm includes momentum correction, local gradient client, momentum factor masking and warm up training, and the compression ratio of the gradient is from 270\(\times\) to 600\(\times\) without loss of accuracy. In the face of Non-IID FL scenario, Felix Sattler et al. (2019)\cite{sattler2019sparse} proposed a new ternary sparse compression framework (STC), which extends Top-k gradient compression and optimal golomb coding to deeply compress the locality. Experiments show that this method can effectively reduce the communication cost. In addition, there are some studies that quantify the sparse gradient on the basis of gradient sparsification\cite{DBLP:conf/iclr/LinHM0D18,alistarh2016quantized} , so that the values of non-zero elements in the updated gradient vector always belong to a preset numerical set, the parameter update vector can subsequently be coded on the basis of parameter update vector sparsity to further reduce the transmission volume.

\subsection{Secure Aggregation}
The secure aggregation framework completes the key exchange through Diffie-Hellman (DH) protocol\cite{DBLP:journals/tit/DiffieH76}. There are public keys between multiple participants. After the local training is completed, the gradient will not be sent directly, but a mask matrix equal to the size of the gradient matrix will be generated according to the public key of DH protocol. By adding the random mask to the original gradient\cite{DBLP:conf/ccs/BellBGL020,DBLP:journals/jsait/SoGA21a,DBLP:journals/corr/abs-2009-11248,DBLP:conf/isit/ZhaoS21}, the real information of the gradient is masked, and the server can not directly obtain the gradient information of the participants. However, the secure aggregation framework also faces the communication bottleneck of traditional distributed training. Due to the need of protecting the privacy of users, the gradient sparsification method can not be directly applied to the security aggregation framework. Generally, there are two ideas to combine gradient sparsification with secure aggregation model, which are respectively to sparse the encrypted gradient and to sparse the gradient before encrypted transmission.

\paragraph{\textbf{Sparse the Encrypted Gradient}}
After local training, each participant calculates the gradient information to be sent, and uses the gradient sparsification method for Top-k selection to record the location of the gradient to be sent. Subsequently, adding the encrypted mask to the original gradient, and then sending only the information at the position corresponding to the record on the masked gradient according to the previous record. In this way, the gradient is protected, and the server cannot collect the real gradient information. However, there is a problem with this method. Even if each participant has a relatively close trend towards the descending direction of the gradient, it can not ensure that all Top-k positions are completely overlapped in each round of training, which will cause that when the participant encrypts a certain part of the gradient vector, Since the other participant holding the symmetric mask is not the corresponding Top-k gradient at this location, the encryption mask at this location will not be sent, so the mask cannot be eliminated in the aggregation process. The mask that cannot be eliminated is equivalent to adding noise to a component position of the gradient vector. If the influence of the mask is too large, it will have a great impact on the overall convergence direction, as well as the convergence speed and the accuracy of the global model. In the case of multi-party participation, this situation may be more serious, so it is not feasible to sparse the encrypted gradient vector directly.

\paragraph{\textbf{Encrypt the Spared Gradient}}
If all positions of the local sparse gradient vector are mask encrypted and sent, this method can ensure that the mask can be eliminated. However, this global sparse gradient encryption scheme violates the original intention of gradient sparsification, because the encrypted mask will cover the communication cost reduced by gradient sparsification. If we want to achieve this level of encryption, the communication cost will be very huge. If the encrypted gradients are float numbers, the number of encrypted gradients sent by each participant is actually no different from the traditional security aggregation framework. At the same time, due to the gradient sparsification method in the local calculation process, this scheme will undoubtedly increase the training time compared with the original scheme, which belongs to the negative optimization scheme. \\\

Compared to the previous work, our proposed method obtain a computation and communication efficient secure aggregation without compromising accuracy.


\section{Efficient and Secure Federated Learning Based on Gradient and Mask Sparsification}
In this section, we concentrates on the detailed designs to these two sparsification methods. The first section, we propose a time-varying hierarchical sparse method for model parameter update, which allows participants only need to send little but critical update when communicate with server, which can greatly reduce the cost of a single communication. On the other hand, we combine the characteristics of gradient sparsification and secure aggregation framework, we make a joint design to ensure that the effect of gradient sparsification on communication optimization is not masked by the encryption mask matrix.

\subsection{Time-varying Hierarchical Gradient Sparsification}

The hierarchical gradient sparsification of the network can reduce the loss caused by the sparsification process. At the same time, considering that in the early stages of training, the network parameters generally vary greatly. After a certain number of update iterations, the amplitude of changes will decrease. Based on this feature, we propose a time-varying hierarchical gradient sparsification (THGS) algorithm. Further improvements are made on the basis of hierarchical sparsification, so that the sparse rate decreases with the increase of iteration rounds, and finally decreases to a set lower bound, as shown in Algorithm 1. We assume that the initial sparsity rate is \(s_{0}\), \(\alpha\) is a constant attenuation factor, and the lower limit of the sparsity rate is \(s_{min}\). In the deep neural network, the sparsity rate \( s_{i}\) of the i-th layer parameter is:
\begin{equation}
s_{i}=\left\{
\begin{aligned}
& s_{0}, & i=1 \\
& s_{i-1} \cdot \alpha, & if\quad\!\!s_{i-1} \cdot \alpha > s_{min}\\
& s_{min}, & else
\end{aligned}
\right.
\end{equation}
\(g_{i}\) is the high-dimensional tensor of the i-th layer parameter, \(L\) is the number of network layers. We use the Top-k selection method to obtain the k-th largest value in \(g_{i}\) and set it as the threshold \(\delta\). Subsequently, we set 1 in \(g_{i}\) which is bigger than the threshold and the rest to 0 to obtain the sparse mask \(\widetilde{g_{i}}\). {The sparse network parameter tensor \(w_{sparse}\) is the hadamard product of \(\widetilde{g_{i}}\) and \(g_{i}\) is the sparse parameters of i-th layer. Then, the client uploards the sparsity update \(w_{sparse}\), and the remaining network parameter tensor residuals \(w_{residual}\) are accumulated.}

\SetCommentSty{mycommfont}

\SetKwInput{KwInput}{Input}                
\SetKwInput{KwOutput}{Output}              

\begin{algorithm}[htpb]
\DontPrintSemicolon
  \label{Algorithm 1}
  \KwInput{\(w, L, s_{0}, s_{min}, \alpha, g_{i}\)}
  \KwOutput{\(w_{sparse}, w_{residual}\)}

  $w_{sparse} \leftarrow torch.zeros\_like(w)$; \\
  $w_{residual} \leftarrow torch.zeros\_like(w)$;\\
  \For{($i=1, i \le L, i++$)}
  {
 $ \widetilde{g_{i}} \leftarrow  to\_vec(abs(g_{i})$; \\
 $ k \leftarrow int(len(\widetilde{g_{i}}) \times s_{i})$; \\
 $ \delta \leftarrow TopK(\widetilde{g_{i}}, k) $; \\
  $ zeros \leftarrow torch.zeros\_like(g_{i}) $; \\
  $ ones \leftarrow torch.ones\_like(g_{i}) $;\\
  $ \widetilde{g_{i}} \leftarrow torch.where(\widetilde{g_{i}} \le \delta, zeros, \widetilde{g_{i}}) $; \\
   $ \widetilde{g_{i}} \leftarrow torch.where(\widetilde{g_{i}} > \delta, ones, \widetilde{g_{i}}) $;\\
   $g_{i-sparse}  \leftarrow  \widetilde{g_{i}} \odot g_{i} $; \\
   $ g_{i-residual} \leftarrow g_{i} - g_{i-sparse} $; \\
   $   w_{sparse} \leftarrow  g_{i-sparse} $;\\
   $w_{residual} \leftarrow g_{i-residual} $;
   }
\caption{THGS algorithm}
\end{algorithm}

The Sparsification is also conducive to improving the security of gradient update of FL. The main implementation means of existing FL gradient attack is that after the server obtains the gradient data of the client, because it has the model starting data before the client calculates the gradient update. It can take the simulated sample data as the learning object and the update gradient actually sent by the client as the label. Taking the difference between the update gradient after the simulated sample data sent into the initial model training and the update gradient actually sent by the client as the penalty function. After that, continuously optimize the simulated sample data so that the gradient calculated by the simulated sample data and the gradient sent by the client reach a very close degree. Then the simulated sample will be close to the local sample data of the client, and the approximate value of the original sample of the client will be obtained. The gradient update only uploads one percent or one thousandth of the real gradient data every time. For the server that wants to carry out gradient attack, the label itself as a penalty function is incomplete data. The simulated sample fitted by this will be far from the real sample, and the ability of the server to carry out gradient attack will be greatly weakened.

\subsection{Sparse the Encryption Mask for Secure Aggregation}
The purpose of applying the gradient sparsification method to the secure aggregation framework is to reduce the cost of communication by sending gradients in important directions at one time while protecting user data privacy. Whether the secure aggregation framework is effective depends on the following two conditions:
\begin{enumerate}
\item After aggregating the encrypted gradient updates of all participants, the server needs to ensure that the masks added by the participants for local data security can be eliminated in the final aggregation result;
\item Compared with the secure aggregation framework without gradient sparsification, the amount of data transmitted during communication is significantly reduced.
\end{enumerate}

In order to ensure that the effect of gradient sparsification on communication optimization is not concealed by the encryption mask matrix, we propose a calculation method for the encryption mask matrix with zero local value, as shown in Algorithm 2.
The participants take the key of DH protocol as a random seed to generate a uniformly distributed mask matrix \(mask_{r}\) \(\in \) \([p, p+q)\). Two mask matrixes are equal if they are both corresponding to the participants and hold the same key. For any participants in the training, the dynamic threshold can be determined according to the total amount of model gradient update \(N\) and the sparsity rate \(R\).
\begin{equation}
R = (\alpha + \beta - \frac{t}{T} ) \cdot R
\end{equation}
where \(\alpha\) is the constant attenuation factor, \(\beta\) is the loss change rate of the participant. The greater the loss change, the more severe the change. \(t\) is the number of iterations, \(T\) is the specified number of training rounds. The more iterations, the smaller the change. \(R\) is the gradient sparsity rate, the upper limit is 1, and the lower limit is the specified minimum sparsity ratio \(R_{min}\).

According to the threshold, we can filter out the mask matrix of Top-k gradient update. {If \(G\) }is the original local gradient update, there is:
\begin{equation}
 {\forall} g_{i} \in G , mask_{top} = \vert g_{i} \vert \geq \sigma
\end{equation}

\(\sigma\) is the random encryption mask filtering threshold:
\begin{equation}
\sigma = p + \frac{k}{x} \cdot q
\end{equation}
where \(x\) is the number of participants and \(k\) is the random mask ratio.

$mask_{e}$ is the random uniform distribution mask matrix filtered by $\sigma$, and $mask_{t}$ is the mask matrix for sparse transmission. Therefore, in order to ensure the safety of the original gradient after sparsification, the gradient after sparsification is updated as:
\begin{equation}
G_{sparse} = encode((G + mask_{e}) \odot  mask_{t} )
\end{equation}

\SetCommentSty{mycommfont}
\SetKwInput{KwInput}{Input}                
\SetKwInput{KwOutput}{Output}              

\begin{algorithm}[htpb]
\DontPrintSemicolon
  \label{Algorithm 2}
  \KwInput{\(T\), dataset \(X\), minibatch size \(b\) per node, the number of nodes \(N\), sparsity rate \(R_{min}\), threshold \(\sigma\), \(\alpha\) ,init parameters \(w = {w[0], w[1], ..., w[M]}\)}
  \KwOutput{$G_{sparse}$}

\For{$t =0, t \le T, t++$}
{
    $G_{t}^{k} \leftarrow G_{t-1}^{k}$;\\

    \For {batch $b \in B$}
    {
       Sample data $x$ from $X$; \\
       $G_{t}^{k}\leftarrow G_{t}^{k} + {1}{Nb}\bigtriangledown f(x; w_{t})$; \\
    }
    $mask_{r} \leftarrow$  generate a mask matrix for DH protocol; \\
    $loss_{0} \leftarrow$  calculate loss value of local model;\\
    $\beta \leftarrow \frac{loss_{0} - loss_{k}}{ loss_{k} }$; \\
    $R_{k} \leftarrow \lbrace R_{k} , R_{min} \rbrace$; \\
    \For{$i=1, i \le M, i++$}
    {
        Select threshold: $\sigma \leftarrow R_{k} of \vert G_{t}^{k}[i]\vert$; \\
        $ mask_{top}[i]  \leftarrow \vert G_{t}^{k}[i] \vert > \sigma$; \\
        \For{$j=1, j \le sizeof  G^k_t[i], j++$}
        {

$mas{k_e}\left[ i \right]\left[ j \right] = \left\{ {\begin{array}{*{20}{c}}
0&{otherwise}\\
{mas{k_r}\left[ i \right]\left[ j \right]}&{mas{k_r}\left[ i \right]\left[ j \right] < \sigma }
\end{array}} \right.$

$mas{k_t}\left[ i \right]\left[ j \right] = \left\{ {\begin{array}{*{20}{c}}
0&{otherwise}\\
1&{mas{k_{top}}\left[ i \right]\left[ j \right] = 0 \wedge  < mas{k_e}\left[ i \right]\left[ j \right] = 0}
\end{array}} \right.$
        }
    }
    $G_{sparse} \leftarrow encode((G_{t}^{k} + mask_{e}) \odot   mask_{t} )$\\
    $G_{residual} \leftarrow G_{t}^{k} \odot  \lnot mask_{t}$ \\
    $loss_{k} \leftarrow loss_{0}$ \\
}

\caption{Secure Aggregation with Mask Sparsification on node k}
\end{algorithm}

The encryption mask matrix with zero local value is to sparse the encryption mask matrix, so that the encryption mask matrix also has a certain proportion of zero. In this way, the existence of a certain proportion of zero can make the communication cost not too large, thereby ensuring the effect of sparseness on communication optimization. In the original secure aggregation framework, there is no need to deliberately design a mask matrix with zero local value, but the mask matrix with zero local value is of great significance in the gradient sparseness of the security aggregation framework. If a certain position of the mask matrix is zero, and the gradient update of the corresponding position does not need to be transmitted in this round, then when the encrypted gradient update data of the participants is aggregated at the server, there is no problem that the mask cannot be eliminated after encryption. At the same time, since the gradient update value of the corresponding position is not transmitted in the actual communication process, the amount of communication data is reduced, and the communication efficiency will also be improved.

\section{Safety Analysis}
The secure aggregation framework is a multi-party encryption computing framework. From the local gradient update encryption process for any participant, it can be inferred that the more participants, the more encryption masks, the more complex the composition of the encryption gradient update, and the more difficult it is for the aggregation end to judge. The local gradient updates the ground truth. It can be seen that when only two parties participate, the encryption situation is simpler, and the composition of the encrypted gradient update obtained by the aggregation end is simpler. If the original gradient mask has an encryption effect, when multiple parties participate, the encryption effect of the local gradient update of the participating parties can also be guaranteed.

In the sparse security aggregation training scenario, the gradient update amplitude does not meet the threshold and the mask component of the corresponding position is zero. In this case, it will not be sent. If the encrypted gradient update meets the transmission conditions, there are several situations: the gradient update amplitude meets the threshold but the corresponding encryption mask is zero, the gradient update amplitude does not meet the threshold but the corresponding encryption mask is not zero, and the gradient update amplitude meets the threshold and the corresponding encryption mask is not zero. Three cases are discussed below:

\begin{enumerate}

\item \textbf{The gradient update magnitude meets the threshold but the corresponding encryption mask is zero.} This situation is equivalent to transmitting the original gradient update information. In the aggregation stage, due to the symmetry of the encryption mask matrix, if another participant happens to select the original gradient update information at the same position, the aggregation cannot infer whether the encryption mask at this position is zero during this round of aggregation. However, as the number of iterations increases, once the corresponding position is zero, the aggregation can determine that the corresponding encryption mask is zero, and reverse the previous gradient data. The true value of the original gradient at the corresponding location is exposed, although the gradient value does not represent the true data. Therefore, the aggregation knows that the mask of some positions is zero, which is not enough to expose the local data. Whether the encrypted information of the mask matrix can be deduced according to the encrypted gradient of the training process, and then the original gradient information can be deduced inversely. The key to continuing to keep local data safe.

\item \textbf{The magnitude of the gradient update does not meet the threshold but the corresponding encrypted mask component is not zero.} This situation is equivalent to transmitting random mask information. In the aggregation stage, according to the symmetry of the encryption mask matrix, there are two situations for the corresponding position: 1. Another participant just selects the original gradient update information and the encryption result of the random mask at the same position. 2. The absolute values of the encrypted values of the two participants are equal and the signs are opposite. For the first case, as long as both positions have non-zero values, the aggregation cannot directly determine whether the encryption mask is zero, and cannot judge the value of the original gradient update. However, once the second situation where the corresponding position is the opposite number occurs, the encrypted mask information will be directly exposed. According to the exposed mask information, the aggregation can easily calculate the gradient of the corresponding position of the participant during the whole training process to update the original value.

\item \textbf{The gradient update magnitude meets the threshold and the corresponding encrypted mask component is not zero.} This situation is similar to the second one. Since the corresponding position is the part of the random mask matrix that is not zero, when at least one of the two positions contains the gradient update value, the value of the corresponding position is not the opposite. , the aggregation cannot directly judge the mask condition. The multi-party training process often requires multiple rounds of iterations. As the iteration progresses, ideally, at least one of the corresponding positions will always have the gradient update range that meets the threshold. However, in the actual training process, the corresponding mask appears to be the opposite number. The situation is very likely to happen. Once a similar mask is exposed, the corresponding position will not have the corresponding encryption effect.
\end{enumerate}

From the above three situations of sending parameters in the gradient sparse scenario, it can be seen that there is a security problem in the local original gradient encryption in the gradient update scenario. First, the local mask of zero will be exposed, revealing part of the original information, but this part The original information is not the complete gradient update information. If the encryption mask can encrypt the components of the local gradient update of other participants, the aggregation still cannot infer the original local gradient update. However, due to the symmetry of the random encryption mask of the secure aggregation framework, in the case of sparse gradients, the gradient information of some positions will not be sent in this round because the magnitude of the gradient update does not meet the threshold. On the contrary, the participants themselves do not send gradients, which does not mean that they do not need to send masks. Since it is not known whether the gradient update of the corresponding position of the other participants is sent during the training process, in order to avoid the error caused by the mask, the participants are randomly masked regardless of the non-zero value. Whether the position corresponding to the code needs to send gradient updates, the random mask information must be sent to avoid random errors after aggregation. However, once the corresponding positions of both parties do not need to send gradient update information, in the encrypted gradient update received by the aggregation, the value of the corresponding position will be the opposite number. In this case, the value of the mask is finally exposed. Subsequent training and previous encrypted gradient updates are ineffective. Except for the case of the opposite number, since the DH protocol is only executed once in this training involving multiple parties, the mask of the corresponding position will not change, even if one party always has a value in the corresponding position, due to the absolute value of the mask. Positive and negative values of equal size frequently appear in corresponding positions, and the aggregation can easily infer the value of the mask by updating the information through multiple rounds of encrypted gradients.

Reviewing the training process of the original secure aggregation framework, the aggregation cannot obtain the original value of encryption mask and local gradient update because only one aggregation result is obtained, and it is not safe to split them. The random mask matrix protects the original value of local gradient update, but accordingly, the local gradient update also protects the mask value from being obtained by the aggregation. The encryption result composed of the two has high security.

Therefore, we use the dynamic sparsity rate method based on the training loss of each participant, which helps to achieve faster convergence. At the same time, because the sparsity rate of each participant is different, and the index of the Top-k parameter will not have a direct impact on the encryption location. A dynamic Top-k gradient parameter helps to protect the original Top-k gradient update. Even if the local Top-k parameters are obtained, the aggregate does not know the total number of Top-k parameters, and the encrypted gradient update with non-zero corresponding positions cannot judge which belongs to the original gradient update. Even in extreme cases, the index and encryption location of Top-k do not overlap at all, but because the aggregation end cannot determine the specific number of Top-k, the original Top-k gradient update cannot be inferred.

\section{Experiments}
To demonstrate the utility of our proposed approaches, we conduct a series of image classification experiments on MNIST, Fashion-MNIST\cite{xiao2017fashion} and CIFAR-10. The MNIST dataset consists of 60,000 training samples and 10,000 test samples. Each sample is a 28 * 28 pixel grayscale handwritten digits. The FMINST is similar to MNIST except that the handwritten digits are replaced by commodity images. The CIFAR-10 dataset contains 60,000 natural images in ten object classes, which consists of 50,000 training pictures and 10,000 test pictures. By using the conventional FL algorithm and the improved algorithm proposed in this paper to train  MNIST-MLP, MNIST-CNN, CIFAR10-CNN, CIFAR10-VGG16 and other models. The goal of our experiments is to demonstrate that with our proposed method, compared to other FL algorithms with no compression, our proposed method achieves high computational and communication efficiency without affecting the accuracy.

Following\cite{mcmahan2017communication}, we train a shared model with 100 total clients, 10 of whom are selected randomly in each round. {The number of local training iterations is 5, and the training batch size is 50.} We use the sample allocation matrix to simulate non-i.i.d. (Independent and identically distributed) training data in the real world and provide different clients with an unbalanced sample from each class. Specifically, in order to simulate the distributed characteristics of the data set in the real scene, the complete training data set needs to be segmented and allocated to the client first in the experiment.

\subsection{{Selection of sparse rate and attenuation factor}}

\begin{figure}
\centering
\includegraphics[width=0.4\textwidth]{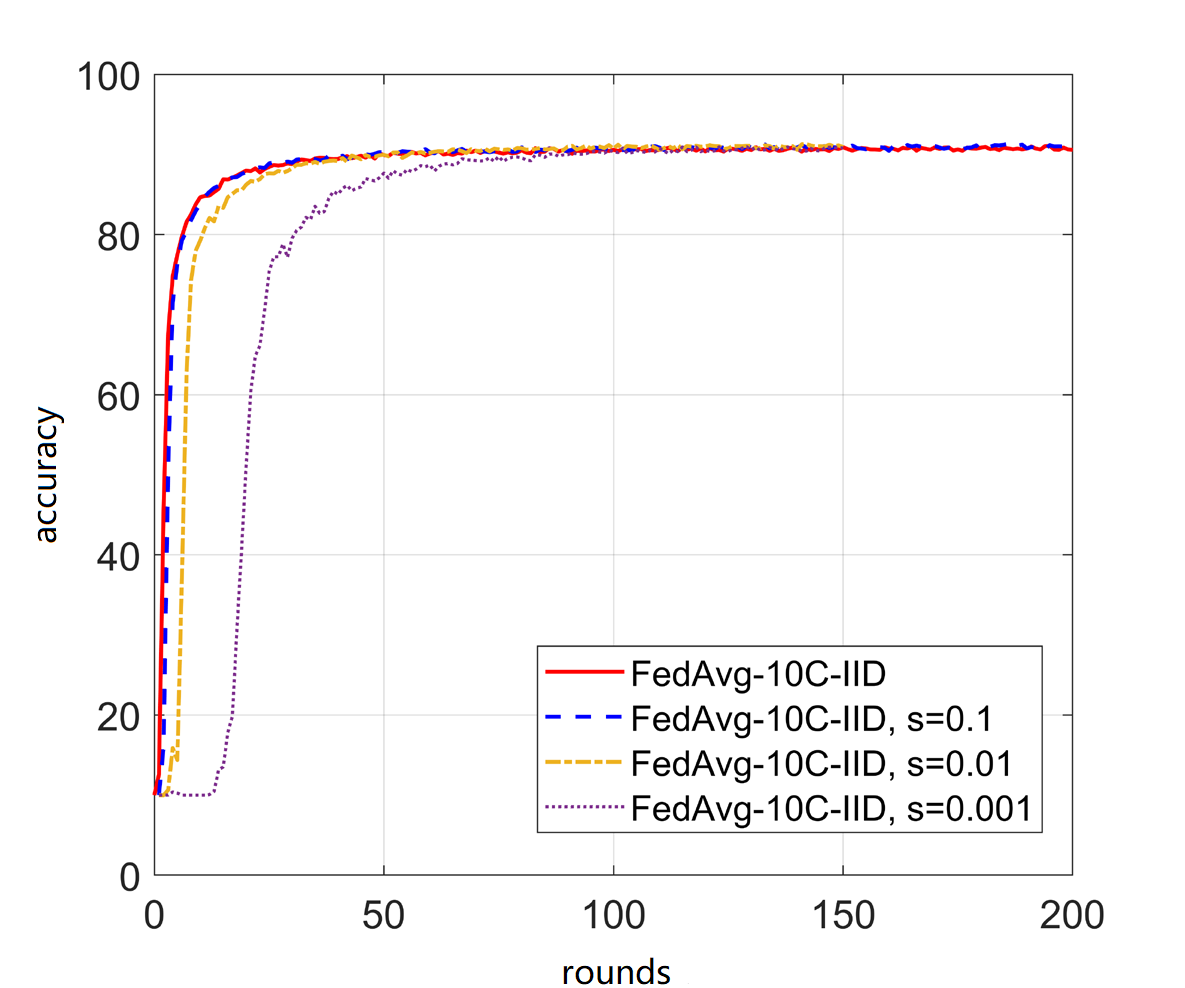}
\caption{The accuracy of the aggregation model updated by gradient sparsification with different sparse rates} \label{Figure 1}
\end{figure}

On the basis of FedAvg, under the IID setting and the experimental configuration, the gradient update operation uploaded by the customer during each iteration is sparsed with different sparse rates, and the experiments with sparse rates \(s\) = 0.1, 0.01, 0.001 are carried out respectively. The experimental results are shown in Figure \ref{Figure 1}.

It can be seen that when the sparsity rate is 0.1, sparsity has almost no effect on the convergence speed of the aggregation model and the final prediction accuracy. When the sparsity rate is further increased to 0.01 and 0.001, the early iterations of the aggregation model will be slowed down. To a certain extent, the performance of the aggregation model will be quickly pulled up after several rounds of iterations.Although the convergence speed itself will be somewhat lower than that of non-sparseness, the predictive ability of the final model will hardly be lost when it converges. The introduction of the algorithm will have a certain negative impact on the performance of the algorithm, which is mainly reflected in a certain reduction in the convergence speed, but this loss is almost negligible compared to the reduction in communication costs caused by sparseness, as shown in Figure \ref{Figure 1}. At a sparsity rate of 0.001, the rounds for the model to converge to the optimal level is about 4 times that of non-sparseness, but the amount of communication required for each round is only one thousand points of the original In contrast, the communication cost can be reduced by hundreds of times. In the same network environment, the time and communication cost required to complete a round of sparse updates are much smaller. Therefore, from the perspective of time, sparseness can speed up the acquisition of models.

\begin{figure*}
\includegraphics[width=0.9\textwidth]{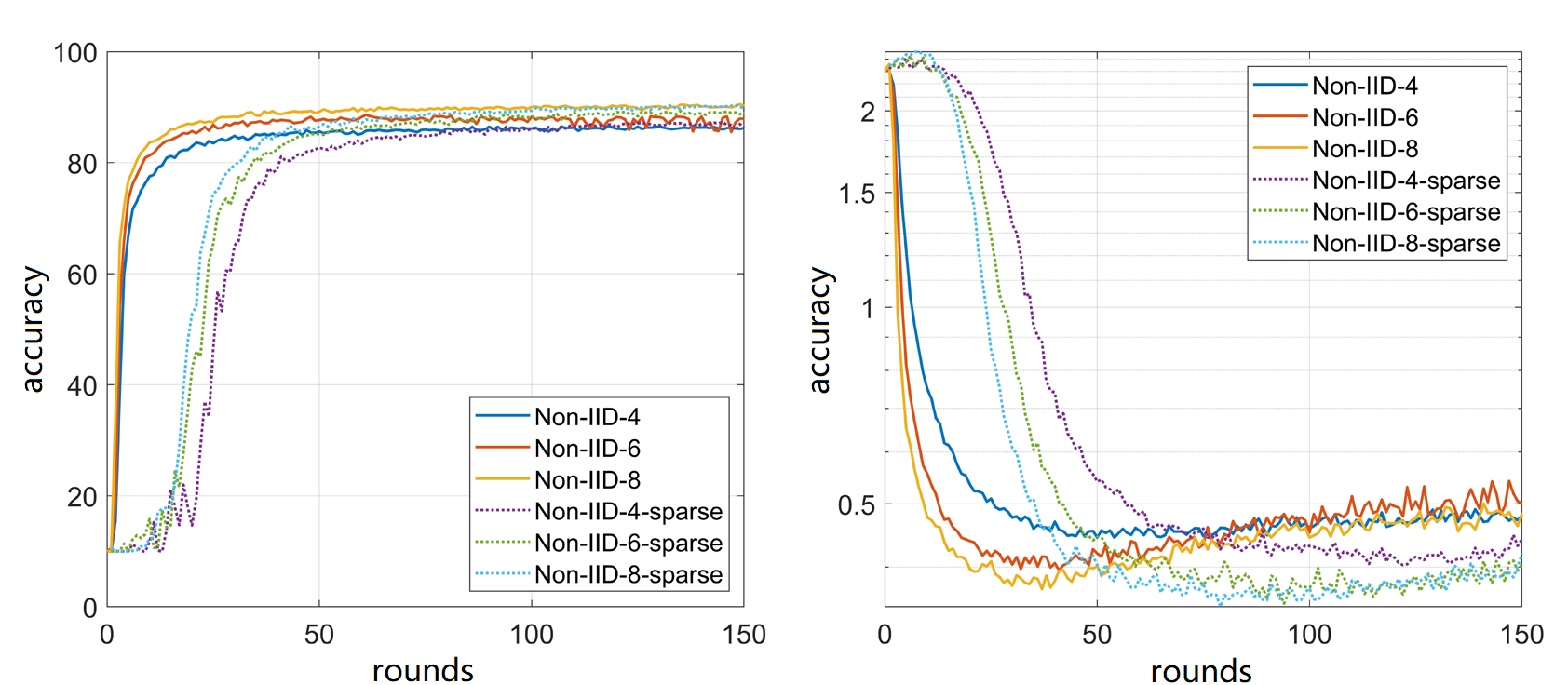}
\caption{Under Non-IID distribution, sparse rate \(s\) = 0.001, the learning curve of the aggregation model under sparse update} \label{Figure 2}
\end{figure*}

For the case of non-IID data set, sparsity is still effective. Take the sparsity rate of \(s\)= 0.001 as an example. As shown in Figure \ref{Figure 2}, through the observation of the loss curve, it can be seen that in some cases, the loss curve after sparse convergence decreases more smoothly than that after non sparse. It is generally believed that the loss curve decreases first and then rises because the model training is sufficient, and further training will bring over fitting effect to the model, The idea of sparseness, which only updates important updates, makes each Federation participant send only the most important updates, which avoids that the parts that may have been fitted in the local training model of each client are aggregated into the aggregation model of the server, and then avoids the over fitting of the model, so that the generalization ability of the aggregation model may be better.

\begin{figure*}[htpb]
\centering
\includegraphics[width=0.75\textwidth]{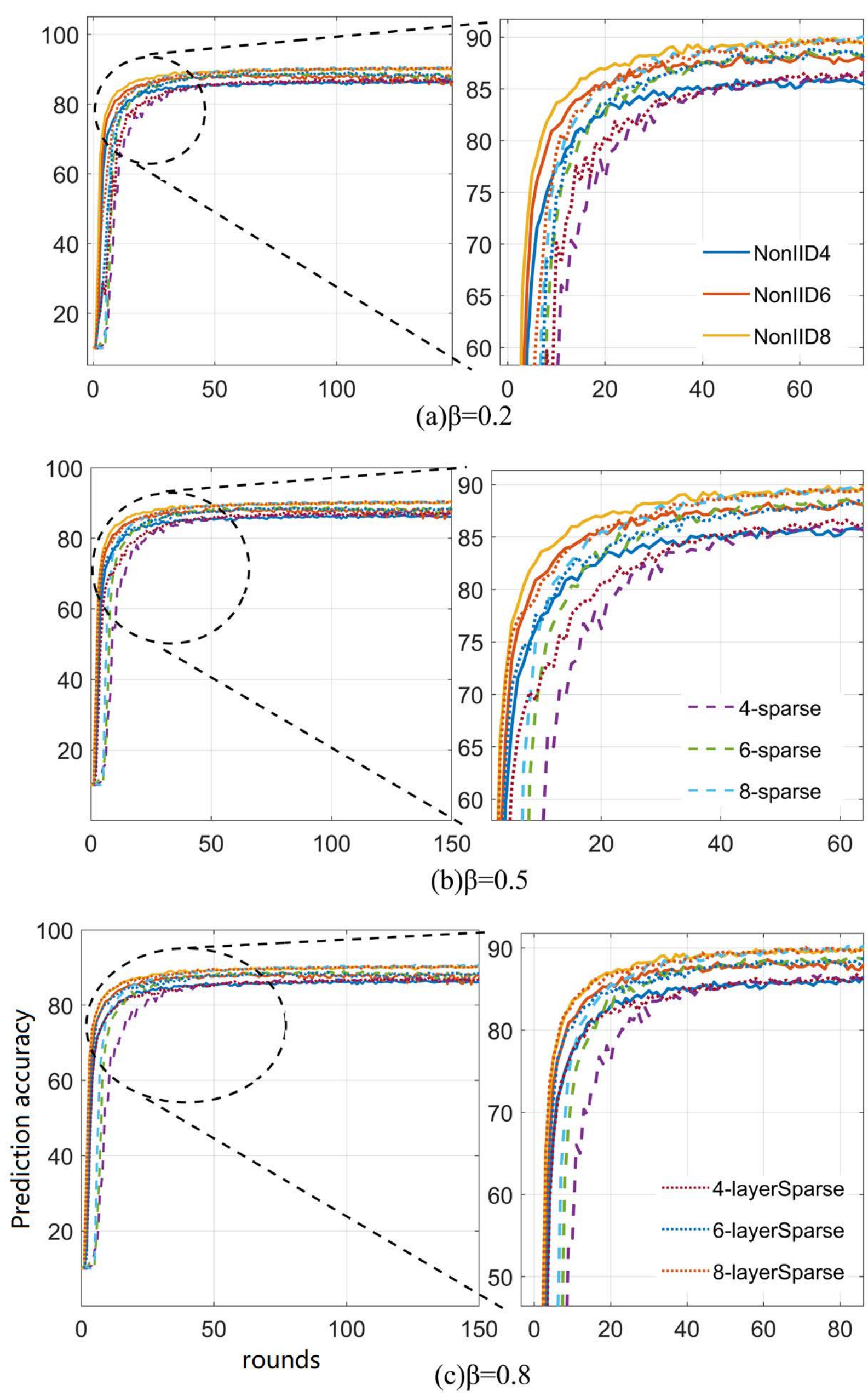}
\caption{Experimental results under Non-IID-4/6/8 distribution with different \(\beta\)} \label{Figure 3}
\end{figure*}

We assume the constant attenuation factor \(\beta\) is 0.2, 0.5 and 0.8 respectively, and the lower limit of the sparsity rate \(s_{min}\) is 0.01. In the iterative process, under these three sparsity rate settings, the tests are carried out under the three conditions of Non-IID-4, Non-IID-6 and Non-IID-8, Non-IID-n (n=1,2,…10) represents a sample with only n types of tags in the client. The results are shown in Figure \ref{Figure 3}. The solid line represents the experimental results of FedAvg algorithm. The long dotted line (- spark) is the experimental results using the conventional sparsification method based on FedAvg, and the short dotted line (- layerspares) is the experimental results using the THGS method proposed in this paper based on FedAvg.

It can be seen that under the three constant attenuation factor, the time-varying hierarchical training effect is better than the conventional sparsity update experimental results. With the improvement of \(\beta\), the effect of the algorithm continues to approach the non-sparsification algorithm, and when the \(\beta\) is 0.8, the loss caused by sparsification can be almost ignored, There is no difference between and non-sparsification updates on the learning curve. Based on the experimental results, we can see the optimization effect of THGS method.

\subsection{Communication cost}

Assuming that the total parameters of the model are \(m\), only the space required to store its gradient vector is calculated, and the subsequent operations such as coding and compression are not considered. Assuming that each parameter is stored in a double-precision floating point, the space required for storing a non sparse gradient vector update is \(m\cdot 64bit\).

For a sparse gradient update vector with a sparsity rate of \(s\), since there are a large number of \(0 \) elements in the vector, it is not necessary to store these vector vacancies. Only the position index and corresponding value of the non-zero elements in the vector need to be stored. Therefore, the space required to store them is:

\begin{equation}
m \cdot s \cdot 64bit + m \cdot  s \cdot 32bit = ms \cdot 96bit
\end{equation}
where \(64bit\) represents the storage space of adouble-precision floating point, and \(32bit\) represents the position index of the non-zero element in the storage sparse vector in the entire vector. Then for the overall process of FL, the total communication overhead required to train a model is:

\begin{equation}
\begin{aligned}
c_{percent} = n_{percent} \cdot ((C\cdot K)\cdot (c_{up}+c_{down}))
\end{aligned}
\end{equation}
where \( n_{percent}\)  is the aggregation rounds required when the prediction accuracy of the model reaches the target convergence accuracy, \(C\cdot K\) is the number of clients selected in each aggregation iteration, \(c_{up}\) and \(c_{down}\) represent the communication cost required for a single client to upload and  download a gradient update respectively:

\begin{equation}
\left\{
\begin{aligned}
& c_{up} = \left\{
                    \begin{aligned}
                    & m \cdot s \cdot 96bit, & if  sparse \\
                    & m \cdot 64bit, & else
                    \end{aligned}
                    \right. \\
& c_{down} = m \cdot 64bit
\end{aligned}
\right.
\end{equation}

For the models used in the experiment, the parameter volumes of different models can be obtained according to the above calculation method. The specific data are shown in Table \ref{Table1}.

\begin{table*}[htpb]
\caption{Different model parameter size and update volumes.\label{Table1}}
\newcolumntype{C}{>{\centering\arraybackslash}X}
       \begin{tabular*}{\hsize}{@{}@{\extracolsep{\fill}}ccccccc@{}}
    \toprule
          & \multicolumn{2}{c}{MNINST} & \multicolumn{2}{c}{Fashion-MNIST} & \multicolumn{2}{c}{CIFAR-10} \\
\cmidrule{2-7}          & MLP   & CNN   & MLP   & CNN   & MLP   & VGG16 \\
    \midrule
    parameter size & \multicolumn{1}{r}{159010} & \multicolumn{1}{r}{582026} & \multicolumn{1}{r}{159010} & \multicolumn{1}{r}{582026} & \multicolumn{1}{r}{5852170} & \multicolumn{1}{r}{14728266} \\
    update volume & 1.2M  & 4.44M & 1.2M  & 4.44M & 44.6M & 112M \\
    \bottomrule
    \end{tabular*}%
\end{table*}

According to the aggregation rounds required for convergence, the communication cost required for different algorithms to complete a FL under different experimental conditions can be calculated. For simplicity, the upload communication cost required to make the aggregation model reach 95\% of the final average convergence accuracy under the Non-IID setting is calculated here. The data are shown in Table  \ref{Table2}.

\begin{table*}[htpb]
\caption{Under Non-IID distribution, the upload communication cost required to reach 95\% of the accuracy when the final average convergence is achieved.\label{Table2}}
\newcolumntype{C}{>{\centering\arraybackslash}X}
    \begin{tabular*}{\hsize}{@{}@{\extracolsep{\fill}}ccccccc@{}}
    \toprule
          & \multicolumn{2}{c}{MNINST} & \multicolumn{2}{c}{Fashion-MNIST} & \multicolumn{2}{c}{CIFAR-10} \\
\cmidrule{2-7}          & MLP   & CNN   & MLP   & CNN   & MLP   & VGG16 \\
\cmidrule{2-7}    \multicolumn{1}{c}{\multirow{2}[1]{*}{FedAvg}} & 840M  & 1332M & 432M  & 4.68G & 94.5G & 156G \\
          & \multicolumn{1}{r}{$\times$13.6} & \multicolumn{1}{r}{$\times$6.11} & \multicolumn{1}{r}{$\times$7} & \multicolumn{1}{r}{$\times$19.8} & \multicolumn{1}{r}{$\times$34} & \multicolumn{1}{r}{$\times$24.6} \\
    \multicolumn{1}{c}{\multirow{2}[0]{*}{FedProx}} & 444M  & 1154M & 552M  & 5.78G & 77.5G & 128G \\
          & \multicolumn{1}{r}{$\times$7} & \multicolumn{1}{r}{$\times$5.3} & \multicolumn{1}{r}{$\times$9} & \multicolumn{1}{r}{$\times$24.5} & \multicolumn{1}{r}{$\times$28} & \multicolumn{1}{r}{$\times$20.2} \\
    Ours  & 61.8M & 218M  & 61M   & 242M  & 2.77G & 6.33G \\
    \bottomrule
    \end{tabular*}%
\end{table*}

It can be seen that the compression amount of uploaded data mainly comes from the gradient and mask sparsification proposed in this paper, which provides {5.3} to 34 times the compression amount. Considering that in the actual scenario, the upload bandwidth of the device is generally far less than the download bandwidth, the algorithm in this paper can reduce the upload communication overhead by dozens of times at the highest. This optimization is considerable.

\section{Conclusion}
In this paper, We have carried out targeted research on the application of sparsity in federated learning and propose an efficient and secure FL framework which can greatly reduce the communication cost of a single communication on the premise of ensuring privacy. At the same time, it has less model performance loss than the traditional sparsity method. Experiments show that under different Non-IID experimental settings, the proposed algorithm can reduce the upload communication cost to about 2.9\% to 18.9\% of the conventional FL algorithm when the sparsity rate is 0.01.

Future work can also consider adding gradient correction and batch normalized update and local gradient to the sparse gradient update process, so as to maintain model accuracy after high ratio sparsification. And adaptive sparsity is used to automatically control the trade-off between optimal communication and computation. In terms of security aggregation, DH key exchange before each round of training undoubtedly increases the training time. Although the sparse mask matrix is conducive to speeding up the training speed, the extra DH key exchange time will undoubtedly affect the final effect of communication optimization. This part needs to be further studied.

\bibliographystyle{IEEEtrans}
\bibliography{reference}

\end{document}